\documentclass{article}
\usepackage[dvipsnames,table]{xcolor}

\usepackage{iclr2025_conference,times}

\usepackage{amsmath,amsfonts,bm}

\def\eqref#1{equation~\ref{#1}}
\def\1{\bm{1}}

\DeclareMathAlphabet{\mathsfit}{\encodingdefault}{\sfdefault}{m}{sl}
\SetMathAlphabet{\mathsfit}{bold}{\encodingdefault}{\sfdefault}{bx}{n}

\DeclareMathOperator*{\argmax}{arg\,max}

\usepackage{wrapfig}
\usepackage{lipsum}
\usepackage{array}
\usepackage{multirow}
\usepackage{hyperref}
\usepackage{url}
\usepackage{graphicx}
\usepackage{algorithm}
\usepackage{amsmath}
\usepackage{algpseudocode}
\usepackage{booktabs}
\usepackage{pifont}
\usepackage{listings}
\usepackage{tcolorbox}

\title{\method: Learning to Reason Dynamically in LLMs via Optimal Reasoning Trajectories Search}

\author{Murong Yue$^\alpha$\thanks{Work done during Murong Yue's internship at Tencent AI Lab.}\hspace{4mm}Wenlin Yao$^\beta$\hspace{4mm} Haitao Mi$^\beta$\hspace{4mm} Dian Yu$^\beta$\hspace{4mm} Ziyu Yao$^\alpha$\hspace{4mm} Dong Yu$^\beta$ \\
$^\alpha$George Mason University\\
$^\beta$Tencent AI Lab, Bellevue\\
\texttt{\{myue,ziyuyao\}@gmu.edu}\\ \texttt{\{wenlinyao,haitaomi,yudian,dyu\}@global.tencent.com}
}

\usepackage{xspace}

\newcommand{\method}{\textsc{Dots}\xspace}

\definecolor{TableGreen}{RGB}{0, 196, 0 }
\newcommand{\tc}[1]{\textcolor{TableGreen}{#1}}

\iclrfinalcopy
\begin{document}

\maketitle

\begin{abstract}
Enhancing the capability of large language models (LLMs) in reasoning has gained significant attention in recent years. Previous studies have demonstrated the effectiveness of various prompting strategies in aiding LLMs in reasoning (called ``reasoning actions''), such as step-by-step thinking, reflecting before answering, solving with programs, and their combinations. However, these approaches often applied static, predefined reasoning actions uniformly to all questions, without considering the specific characteristics of each question or the capability of the task-solving LLM. In this paper, we propose 
\textbf{\method}, an approach enabling LLMs to reason \underline{D}ynamically via \underline{O}ptimal reasoning \underline{T}rajectories \underline{S}earch, 
tailored to the specific characteristics of each question and the inherent capability of the task-solving LLM. 
Our approach involves three key steps: i) defining atomic reasoning action modules that can be composed into various reasoning action trajectories; ii) searching for the optimal action trajectory for each training question through iterative exploration and evaluation for the specific task-solving LLM; and iii) using the collected optimal trajectories to train an LLM to plan for the reasoning trajectories of unseen questions. In particular, we propose two learning paradigms, i.e., fine-tuning an external LLM as a planner to guide the task-solving LLM, or directly fine-tuning the task-solving LLM with an internalized capability for reasoning actions planning.
Our experiments across eight reasoning tasks show that our method consistently outperforms static reasoning techniques and the vanilla instruction tuning approach. Further analysis reveals that our method enables LLMs to adjust their computation based on problem complexity, allocating deeper thinking and reasoning to harder problems. Our code is available at \url{https://github.com/MurongYue/DOTS}.
\end{abstract}
\vspace{-1.5em}
\section{Introduction}
% Reasoning ability is a crucial component of human cognition.
\vspace{-1mm}
Large Language Models (LLMs) have demonstrated remarkable performance in solving complex reasoning tasks~\citep{rae2021scaling, lewkowycz2022solving, zhong2023agieval}, such as math reasoning~\citep{imani2023mathprompter,ahn2024large}, symbolic reasoning~\citep{kojima2022large}, and common-sense reasoning~\citep{krause2023commonsense,zhao2024large}. 
% However, their ability is still limited, and recent studies have extensively explored how to bridge the gap.
The dominant approaches to eliciting reasoning capability in LLMs mainly fall into two categories, i.e., instruction tuning and prompt engineering.
% The main approaches to enhance LLMs' reasoning ability are instruction tuning~\citep{wang2022benchmarking} and prompting engineering~\citep{sahoo2024systematic}. 
Instruction tuning~\citep{wang2022benchmarking} collects question-answer pairs about the reasoning task and employs supervised fine-tuning to optimize an LLM for better reasoning performance~\citep{yue2024mammoth2,tang2024mathscale}, with recent effort focusing on improving the scale and the quality of the fine-tuning data~\citep{luo2023wizardmath,peng2023instruction,yue2023mammoth, yue2024mammoth2,chan2024scaling}. 
% The pairs are collected via human annotation~\citep{wang2022benchmarking} or prompting LLMs such as GPT-4 with seed data~\citep{luo2023wizardmath,peng2023instruction}.
Prompt engineering instead aims to design better prompts to elicit the reasoning capability of an LLM without updating its parameters.
% provide expert-designed instructions to guide LLMs in performing specific reasoning actions. 
The Chain-of-Thought (CoT) approach~\citep{wei2022chain,kojima2022large} prompts an LLM to answer the reasoning question step by step in natural language, and program-aided approaches~\citep{chen2022program,gao2023pal} prompt the LLM to write executable code and leverage an interpreter to execute code for obtaining the final result. Besides, prompting the LLM to decompose the question before answering it~\citep{radhakrishnan2023question,zhou2023leasttomost}, or to verify the solution before returning it as the final answer~\citep{madaan2024self}, has also been proven effective in specific reasoning tasks.

However, both types of approaches suffer from a critical limitation, i.e., being unable to \emph{dynamically decide the best reasoning strategies}. For instruction-tuning-based approaches, the fine-tuned LLMs are constrained to follow the same reasoning format of the training data (e.g., CoT~\citep{luo2023wizardmath}) and lack the flexibility to adopt other reasoning strategies. 
% This has shown to be an issue even with the recent OpenAI o1-preview model; for example, a simple ``Hello'' message could trigger the unnecessary reasoning process of the o1-preview model, causing it to take multiple seconds to give a response.
% This has shown to be an issue even with GPT-4o; for example, when presented the question\footnote{A book with 50 pages numbered 1 through 50 has its pages renumbered in reverse, from 50 to 1. For how many pages do both sets of page numbers share the same one's digit?}, GPT-4o responds with natural language and makes an error, overlooking the potential of the more precise and efficient approach of solving with programs (Appendix~\ref{Appendix: case}). 
An example revealing a similar weakness of GPT-4o is shown in Appendix~\ref{Appendix: case}.
% as shown in Appendix~\ref{Appendix: case}. It overlooks the potential for a more precise and efficient method through code generation.
% \wenlin{This example is very long and a little distractive. Maybe put it into a footnote?}
On the other hand, current prompt engineering approaches assume predefined prompting strategies and uniformly apply the same to every question. However, different types of questions are better suited to different reasoning strategies~\citep{zhao2023cotvspot}, and the effectiveness of a prompting approach also depends on the inherent capability of the task-solving LLM (e.g., LLMs pre-trained on code data are better at programming-aided reasoning). Consequently, the same prompt may not be equally effective for every question and every LLM. 
% Recent approaches are designed to allow for instance-specific prompt optimization~\citep{srivastava-etal-2024-instances}, while they still only focus on optimizing CoT instructions to be less ambiguous, failing to make LLMs leverage different reasoning actions.

% \zyc{note that some prompt optimization approaches may allow for question- and model-specific prompt selection. An example is our paper at ACL \cite{srivastava-etal-2024-instances}, although it did not focus on reasoning and did not define structured reasoning actions.}

% \begin{figure*}[t!]
%     \centering
% \includegraphics[width=0.75\textwidth]{figures/intro.pdf}
% \caption{Comparison of the static reasoning actions and our dynamic reasoning actions. EMPTY means we do not take any reasoning action here. The static reasoning actions may not work well for different questions, while our method could adapt to different questions via dynamic reasoning actions.\zyc{self-refine should not be coupled with PoT; they are in two orthogonal dimensions. Spell out "CoT" and cite Wei et al. Cite the self-refine paper.}
%     }
%     \label{fig:intro}
% \end{figure*}
\begin{figure*}[t!]
    \centering
% \vspace{-1em}
\includegraphics[width=\textwidth]{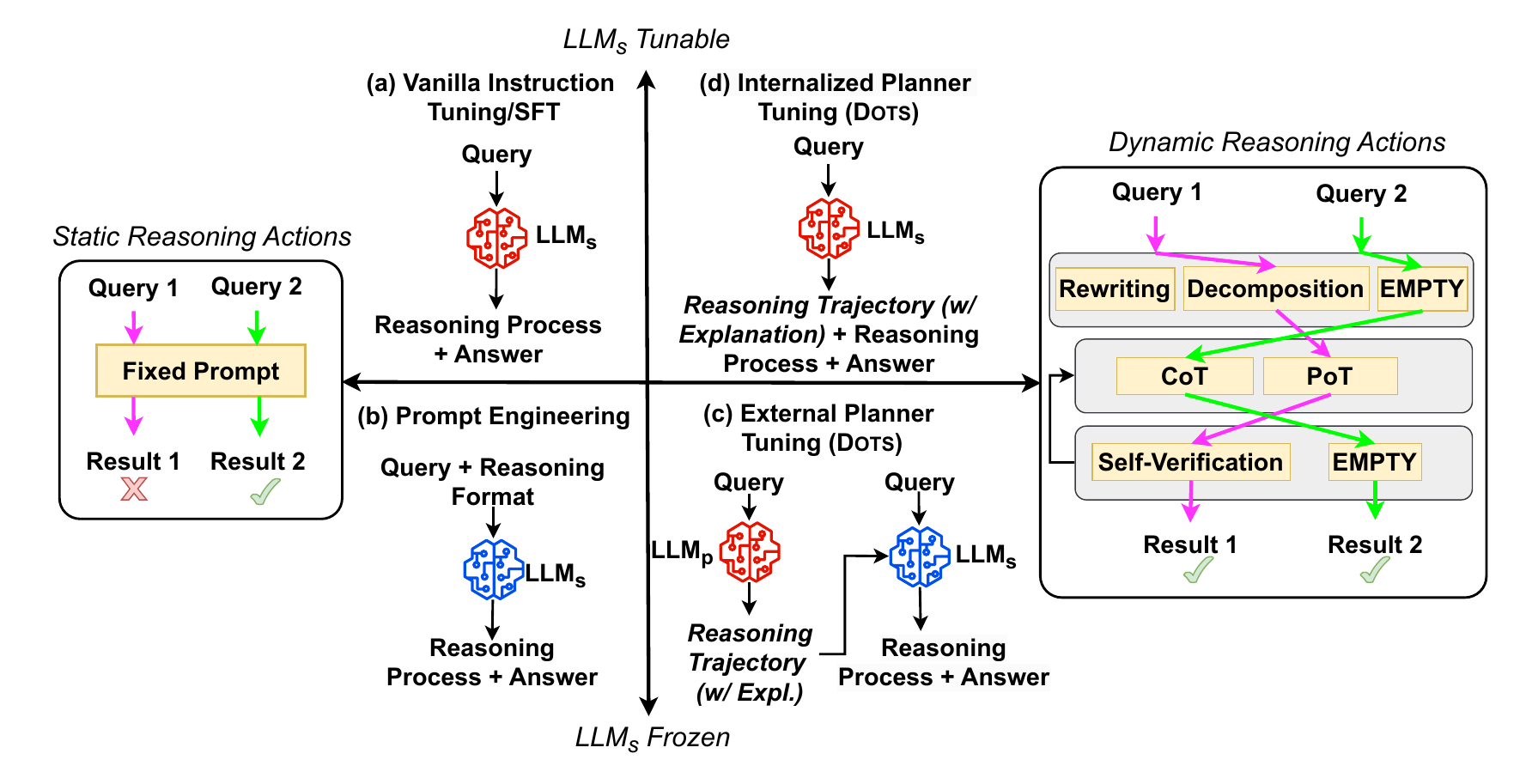}
\vspace{-2em}
\caption{A comparison of different paradigms of LLM reasoning. 
% \wenlin{Can we use a larger text font in this figure?} 
Unlike prior approaches with predefined, static reasoning actions, \method dynamically plans for the optimal reasoning trajectory per each question and the specific task-solving LLM ($LLM_s$). In particular, \method encompasses two inference setups, i.e., external planner tuning (c) and internalized planner tuning (d), depending on whether to introduce an external LLM as a planner ($LLM_p$) or to internalize the trajectory planning capability into the same solver LLM ($LLM_s$). (\includegraphics[width=0.02\textwidth]{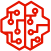}: tunable; \includegraphics[width=0.02\textwidth]{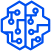}: frozen)}
\vspace{-2em}

    \label{fig:intro}
\end{figure*}
In this paper, we present \method, an approach empowering LLMs to actively select optimal reasoning actions for given questions and the task-solving LLM (Figure~\ref{fig:intro}). 
% As shown in Figure~\ref{fig:intro}, our method empowers LLMs with dynamic reasoning capabilities to be tailored to the characteristics of each question. 
We begin by constructing atomic reasoning action modules, which are composed to generate multiple potential reasoning action trajectories. Then we collect the training data by searching for an optimal (in terms of both its success rate and the number of reasoning actions needed) action trajectory through numerous explorations and evaluations. This optimal trajectory is tailored to the specific task-solving LLM.
Subsequently, we employ supervised fine-tuning to train an LLM in determining the optimal reasoning action trajectory. We implement this approach in two distinct setups: (1) For closed-source or computationally costly task-solving LLMs, we fine-tune a smaller LLM as an external planner to predict optimal reasoning actions for the task-solving LLM; (2) For open-source and small-size LLMs, we fine-tune the task-solving LLM itself to plan on the reasoning actions to take before solving the reasoning task, internalizing the autonomous planning capability directly into the LLM. This dual approach allows for flexible application across different LLM accessibility constraints.

Our experimental results demonstrate the efficacy of our proposed method in enhancing the reasoning capabilities of LLMs. We conducted extensive evaluations across multiple LLMs (GPT-4o-mini, Llama3-70B-Instruct, and Llama3-8B-instruct~\citep{dubey2024llama3}) and a diverse set of reasoning tasks, encompassing in-distribution, few-shot, and out-of-distribution scenarios. The results reveal that \method consistently outperforms static prompt engineering techniques and vanilla instruction tuning methods across various reasoning challenges. Through a comprehensive ablation study, we validate the significance of each component in our methodology.
Moreover, our analysis of reasoning action distributions highlights that our method can adapt to the specific characteristics of reasoning questions and the inherent capability of task-solving LLMs. 
We further confirm that our method incurs minimal additional financial costs. Lastly, we showcase that LLMs can naturally develop the capacity to allocate more computational resources to complex problems through a process of exploration and learning, without explicit guidance.
%  organically

% \section{\method: Dynamic Reasoning Actions Learning}
\vspace{-1em}
\section{\method: Learning to Reason Dynamically}
% \wenlin{The title of this section is the same with our paper title. Maybe change to a shorter one?}
\vspace{-2mm}
\subsection{Overview}
\begin{figure*}[t!]
    \centering
    \vspace{-2mm}

\includegraphics[width=1\textwidth]{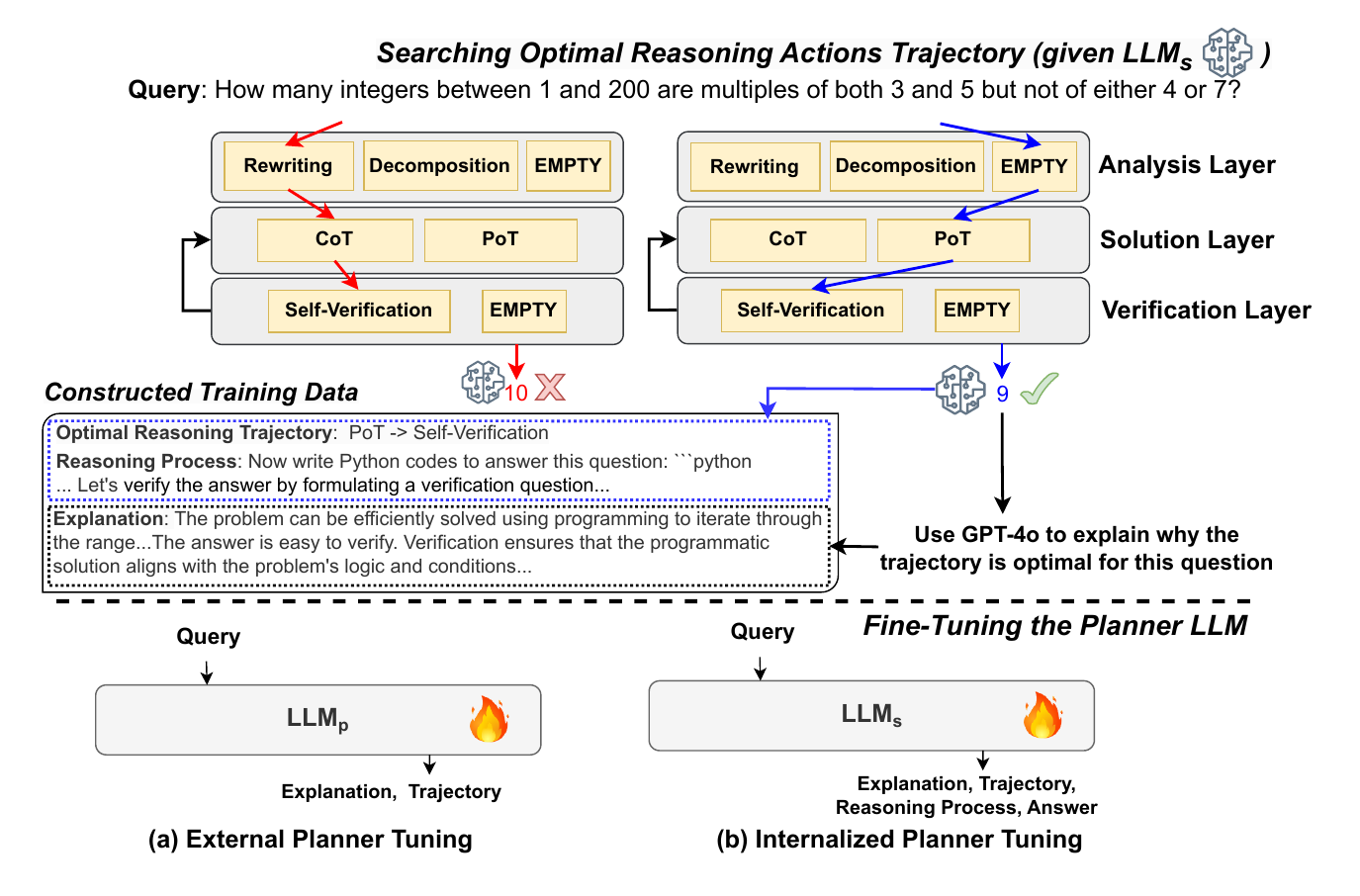}
\vspace{-2em}
    \caption{The training process of \method, including searching for the optimal reasoning trajectories for questions in the training set and fine-tuning the internalized/external planner LLM.
    }
    \vspace{-1em}
    \label{fig:structure}
\end{figure*}

\vspace{-1mm}
Our goal is to enable LLMs to select the most effective reasoning actions autonomously. 
Denote $LLM_{s}$ as the task-solving LLM, $Q$ as the input query, $p$ as the reasoning action trajectory path, $E$ as the explanation for a trajectory, and $R$ as the reasoning process leading to the final answer $y$. Our approach encompasses two setups during the inference stage (Figure~\ref{fig:intro}):
\vspace{-0.5em}
\vspace{-1mm}
\paragraph{External Planner Tuning}
This setup is designed for scenarios where the solver ($LLM_{s}$) is a closed-source LLM or is computationally costly to train. As depicted in Figure~\ref{fig:intro} (c), we train an external planner, denoted as $LLM_{p}$, to determine the optimal reasoning actions:
\begin{equation}\label{eq:external planner step1}
(E,p) =\text{$LLM_{p}$}(Q;\theta_p)
\end{equation}
where $\theta_p$ is the parameters of $LLM_{p}$. We empirically found that training the planner to explain its trajectory selection ($E$) helps its learning.
Upon obtaining reasoning actions, the solver $LLM_{s}$ parameterized by $\theta_s$ then proceeds to generate the reasoning process $R$ and the final answer $y$:
\begin{equation}\label{eq:external planner step2}
(R,y) =\text{$LLM_{s}$}(Q,T; \theta_s) 
\end{equation}
% where $\theta_s$ is the parameters of $LLM_{s}$.
% \vspace{-0.5em}
\vspace{-2em}
\paragraph{Internalized Planner Tuning} 
This setup is designed for task-solving LLMs ($LLM_{s}$) that are open-source and small-size. In this case, we propose to \emph{internalize} the trajectory planning capability into the task-solving LLM by training it to simultaneously learn to plan and learn to perform the reasoning task. 
% we can optimize both the planning ability and the question-solving ability simultaneously. 
As shown in Figure~\ref{fig:intro} (d), the final answer $y$ is obtained by:
\begin{equation}\label{eq:internalized planner}
(E,p,R,y) =\text{$LLM_{s}$}(Q; \theta_s)
\end{equation}
An overview of \method's learning process is presented in Figure~\ref{fig:structure}, consisting of three key steps: (i) \textbf{Defining atomic reasoning modules:} We define several atomic reasoning modules, each representing a distinct reasoning action, (ii) \textbf{Searching for optimal action trajectories:} We conduct explorations and evaluation of various reasoning paths to identify optimal reasoning actions for questions in the training data, and (iii) \textbf{Fine-tuning LLMs to plan for optimal reasoning trajectories:}
We fine-tune LLMs to autonomously plan the reasoning action trajectory under the two aforementioned setups.
In what follows, we elaborate on each step.

% Comments: font is too small.

\begin{table}[htp]
\vspace{-1.5em}
\caption{Prompt engineering methods with different reasoning actions. Our method could dynamically select reasoning actions among all of them.}\vspace{3mm}
\centering
\resizebox{0.9\textwidth}{!}{%
\begin{tabular}{l|cc|cc|c}
\hline
\multirow{2}{*}{\textbf{Prompting Method}} &
  \multicolumn{2}{c|}{\textbf{Analysis Layer}} &
  \multicolumn{2}{c|}{\textbf{Solution Layer}} &
  \multicolumn{1}{c}{\textbf{Verification Layer}} \\
 &
  Rewriting &
  Decomposition &
  NL &
  Program &
  Verification \\ \hline
CoT~\citep{wei2022chain} &\ding{55} &\ding{55} &\tc{\ding{51}} &\ding{55} &\ding{55} \\
PoT~\citep{chen2022program} &\ding{55} &\ding{55} &\ding{55} & \tc{\ding{51}}&\ding{55} \\
LTM~\citep{zhou2023leasttomost} &\tc{\ding{51}} &\ding{55} &\tc{\ding{51}} &\ding{55} &\ding{55} \\
R\&R~\citep{deng2023rephrase} &\ding{55} & \tc{\ding{51}}&\tc{\ding{51}} &\ding{55} &\ding{55} \\
Self-Refine~\citep{madaan2024self} &\ding{55} &\ding{55} &\tc{\ding{51}} &\tc{\ding{51}} & \tc{\ding{51}}\\
Self-Verification~\citep{weng2022selfverification} &\ding{55} &\ding{55} &\tc{\ding{51}} &\ding{55} &\tc{\ding{51}} \\
PromptAgent~\citep{wang2023promptagent} &\tc{\ding{51}} &\tc{\ding{51}} &\tc{\ding{51}} &\ding{55} &\ding{55} \\
\method (\bf ours) &  \tc{\ding{51}} &\tc{\ding{51}} &\tc{\ding{51}} &\tc{\ding{51}} &\tc{\ding{51}} \\

            \bottomrule
\end{tabular}
}
\vspace{-1em}

\label{table: modules}
\end{table}

\subsection{Defining Atomic Reasoning Actions Modules}
\vspace{-1mm}
Prior studies have validated the effectiveness of various reasoning strategies (Table~\ref{table: modules}). We build on top of them and categorize the existing strategies as reasoning actions across three layers:
\vspace{-0.5em}
\vspace{-1mm}
\paragraph{Analysis Layer} 
Actions in this layer enable the LLM to analyze the input query before attempting to solve it, including 
(1) \texttt{Query rewriting}: reformulating the query to enhance comprehension, and (2) \texttt{Query decomposition}: breaking down the initial question into multiple, more manageable sub-questions. We denote the action taken in this layer as $A_{a}$.

\vspace{-0.5em}
\vspace{-1mm}
\paragraph{Solution Layer} 
% The reasoning actions in this layer focus on problem-solving. 
% Different problems require problem-solving actions tailored to their specific characteristics. 
% Some are better suited to step-by-step natural language thinking, while others are more appropriate for writing code in aiding to solve them. Therefore, we involve two methods in this layer: 
Actions in this layer consider variants in the reasoning format. Prior works showed that different queries are better solved following different reasoning processes~\citep{zhao2023cotvspot}. In our work, we consider the most commonly adopted formats, i.e.,
(1) \texttt{CoT}: solving the question step-by-step in natural language, and (2) \texttt{PoT}: addressing the question through code generation. We denote the action chosen in this layer as $A_{t}$. % s is used previously

% \vspace{-0.5em}
\vspace{-1em}
\paragraph{Verification Layer} 
Finally, the verification layer is responsible for checking the correctness of the proposed solution.
It is particularly useful for problems where verification is significantly easier than solving the problem itself, e.g., the Game of 24 \citep{yao2024tree}.
Therefore, we set a \texttt{Self-Verification} action module in this layer.
If this module determines that the reasoning process from the solution layer is incorrect, the LLM will revert to the solution layer to reattempt to solve the problem.
During this reattempt, the LLM is provided with both the initial answer and the feedback from the verifier explaining why the initial answer was incorrect. 
The process continues until the verifier confirms that the answer is correct or the pre-defined maximum number of iterations for self-verification is reached. 
We denote the action taken in this layer as $A_{v}$.
% when the self-verification reaches the allowable maximum iteration.

We observe that introducing excessive reasoning actions can lead to increased latency, and even sometimes result in incorrect answers. To mitigate this issue, we introduce an \texttt{Empty} action in both the analysis and the verification layers, allowing the LLM to bypass these two layers when dealing with simple questions. Detailed prompts for each module are provided in Appendix~\ref{Appendix: Prompts}.
% \vspace{-0.5em}

\vspace{-1.5mm}
\subsection{Searching for Optimal Reasoning Action Trajectories}
\vspace{-1mm}
% \zyc{Need to clarify 1) what LM we use in this searching process; 2) any difference in external vs. internalized planner setting? 3) where does the Q come from?}\murong{solved}
To teach the external/internalized planner to plan for the optimal reasoning trajectory, we start by constructing training data containing questions and their optimal action trajectories for the specific task-solving LLM.  We obtain this by iteratively searching all possible reasoning trajectories for each question, including exploring the current paths and pruning paths that are unlikely to be optimal. The task-solving LLM is used during this search process to generate answers to make the reasoning trajectory align with their intrinsic ability to perform different reasoning actions effectively.

% \begin{algorithm}
% \caption{Searching for the Optimal Reasoning Trajectory}
% \begin{algorithmic}[1]
% \small
% \Require A set of all possible action trajectories $L = \{p_1, p_2, \ldots, p_m\}$, number of iterations $K$, number of evaluations per path $n$, and number of paths to retain in each iteration $retain_1, retain_2,...$.
% \Ensure Optimal action trajectory path $p^*$.
% \State Initialize the set of all possible action trajectories $L$
% \State $L \gets \{(A_a = \text{Empty}, A_s = \text{CoT}, A_v = \text{Empty}) \ldots\}$
% \For{$k = 1$ to $K$}
%     \State Initialize an empty list for success rates $R \gets []$
%     \ForAll{$p \in L$}
%         \State Initialize success rate $r_p \gets 0$
%         \For{$i = 1$ to $n$}
%             \State Execute path $p$ and compute success outcome $s_i \gets \text{execute\_path}(p)$
%             \State Update success rate $r_p \gets r_p + s_i$
%         \EndFor
%         \State Compute average success rate for path $p$: $r_p \gets r_p / n$
%         \State Append the tuple $(p, r_p)$ to $R$
%     \EndFor
%     \State Sort $R$ by:
%         \begin{enumerate}
%             \item Success rate $r_p$ in descending order
%             \item Path length $|p|$ in ascending order (for ties in success rate)
%         \end{enumerate}
%     \State Retain the top $retain_k$ paths with the highest success rates:
%     \State $L \gets \{p_1, p_2, \ldots, p_{retain_k}\}$ from top paths in $R$
% \EndFor
% \State After $K$ iterations, return the optimal path:
% \State $p^* \gets \argmax_{p \in L} r_p$
% \end{algorithmic}
% \label{algorithm: search}
% \end{algorithm}
\begin{algorithm}[t!]
\caption{Searching for the Optimal Reasoning Action Trajectory}
\begin{algorithmic}[1]
\small
\Require Input query $Q$ and ground true answer $y^*$, solver $LLM_s$, max iteration $K$, number of evaluations $N_{eval}$, and number of candidate trajectories to retain in each iteration $N_1, N_2, \cdots, N_K$.
\Ensure Optimal action trajectory path $p^*$ for query $Q$ and solver $LLM_s$.
\State Initialized candidate trajectory set $\mathcal{P} \gets \{(A_a = \text{Empty}, A_t = \text{CoT}, A_v = \text{Empty}), \cdots\}$;
% \State Initialize the iteration record $\mathcal{R}_k \gets [ \; ]$;
\State Initialize the record of the accumulated success rate of each candidate trajectory: $\mathcal{R} \gets \{ p : 0 \mid p \in \mathcal{P} \}$;
\For{iteration $k = 1$ to $K$}
    % \State Initialize the $k$-th iteration record $\mathcal{R}_k \gets \{\}$;
    \ForAll{$p \in \mathcal{P}$}
        \State Execute the trajectory $p$ against $Q$ and $LLM_s$ for $N_{eval}$ times with non-zero temperature and obtain an average success rate $r_p$ (compared to the true answer $y^*$);
        \State Update the accumulated success rate of $p$: $\mathcal{R}[p] \gets \frac{\mathcal{R}[p]\cdot(k-1)\cdot N_{eval} + r_p \cdot N_{eval}}{N_{eval}\cdot k}$;
        % \State Update $\mathcal{R}[p] \gets \left(\frac{\mathcal{R}[p][0]\cdot(k-1)\cdot N_{eval} + r_p \cdot N_{eval}}{N_{eval}\cdot k}, \: \text{trajectory length} \: |p|\right)$;
        % \State $\mathcal{R}_k.\text{append}\big((r_p, \: \text{trajectory length}\: |p|)\big)$.
    \EndFor
    \State Sort $\mathcal{P}$ first by the accumulated success rate $\mathcal{R}[p]$ and then the trajectory length $|p|$ in ascending order;
    % \State Sort $\mathcal{R}_k$ first by each candidate trajectory's $r_p$ and then by its $|p|$;
    \State Reset $\mathcal{P} \gets$ top $N_k$ trajectories in $\mathcal{P}$.
\EndFor
\State \textbf{Return} $p^* \gets \argmax_{p \in \mathcal{P}}\mathcal{R}[p]$.
\end{algorithmic}
\label{algorithm: search}
\end{algorithm}

This searching process is shown in Algorithm~\ref{algorithm: search}. Given the query and ground-truth answer sourced from the training data, the process runs iteratively. In each iteration, the algorithm considers either the full set of candidate trajectories (for iteration $k=1$) or the current best subset (for iteration $k > 1$). Each candidate trajectory is executed for $N_{eval}$ times with a non-zero temperature to obtain a more reliable evaluation of its success rate. We then sort the current subset of trajectories by its success rate accumulated from the past $k$ iterations and then the trajectory length to encourage a shorter trajectory (which is thus computationally more efficient).
Only the top $N_k$ candidates will be retained and rolled over to the next iteration of the assessment. In practice, we opt for a smaller $N_{eval}$ and run the search for multiple iterations, as opposed to finishing the search with a larger $N_{eval}$ in one iteration, as the latter incurs a much larger cost ($N_{eval} \times |\mathcal{P}_0|$ with a large $N_{eval}$ vs. $N_{eval} \times (|\mathcal{P}_0| + N_1+\cdots+N_{K-1})$ with a small $N_{eval}$ in our algorithm).

% \zyc{todo: discard failed examples}\murong{solved}
% In the process of validating various trajectories for each question, we identify cases when \emph{any} trajectory successfully solves the query or all trajectories fail to solve the query and exclude them from the training set, as these do not contribute to the planner LLM's learning of trajectory planning.
In the process of validating various trajectories for each question, we exclude instances where \emph{any} trajectory solves the query or \emph{all} fail to do so, as they do not contribute to the planner LLM's trajectory planning learning.
After identifying the best reasoning trajectory, we leverage GPT-4o to verbally explain why the trajectory is optimal.
Our prompt is shown in Appendix~\ref{Appendix: Prompts}.
This process is applied to all instances in the training data, giving us tuples of query $Q$, ground true answer $y^*$, optimal trajectory $p^*$, and its explanation $E$. 
For internalized planner tuning, we collect the reasoning process $R$ when running the solver $LLM_s$ following the optimal trajectory $p^*$.
% , excluding cases where all action trajectories successfully solve the problem, as these do not contribute to the LLM's learning of action selection. 
\vspace{-1em}
\subsection{Learning to Plan for Optimal Reasoning Trajectories}
\vspace{-1mm}
Having obtained the optimal trajectories, we then use supervised fine-tuning with cross-entropy loss to train the planner LLM to predict optimal trajectories for input questions and the specific solver LLM. 
% We explore two distinct aforementioned setups. 
For external planner tuning, a lightweight $LLM_{p}$ is trained to predict a concatenation of the explanation and the optimal trajectory (Eq~\ref{eq:external planner step1}); for internalized planner tuning, the solver $LLM_s$ is trained to predict the explanation, the optimal trajectory, the reasoning process collected from $LLM_s$ itself, and the true answer $y^*$ (Eq~\ref{eq:internalized planner}).

\vspace{-1em}
\section{Experiment}
\vspace{-2mm}
\subsection{Experimental Setup}
\begin{wraptable}{r}{0.45\textwidth}

\centering
\vspace{-2em}
\caption{Overview of our evaluation datasets. 
% \wenlin{Reasoning Type is not very accurate here. Change to a different name?}
}
\vspace{3mm}
\small
% \vspace{-0.35cm}
\setlength{\tabcolsep}{3pt}
\resizebox{0.45\textwidth}{!}{
\begin{tabular}{l|c|l}
\hline
\bf Dataset       & \bf Distribution    & \bf Task Type \\ 
\hline
MATH          & In Distribution & math         \\
\hline
BBH           & \multirow{3}{*}{Few-shot}  & mixture      \\
Game of 24    &       & numerical    \\
TheoremQA     &       & scientific   \\
\hline
Deepmind Math &  \multirow{4}{*}{Out-of-Distribution}     & math         \\
MMLU-pro      &        & scientific   \\
StrategyQA    &        & common sense \\
DROP          &       & multi-hop  \\
\hline
\end{tabular}
}
\vspace{-1em}

\label{table:dataset}
\end{wraptable}
% \vspace{-0.5em}
\vspace{-1mm}
\paragraph{Datasets}
We evaluate the effectiveness of our method across multiple datasets and various reasoning tasks.
Based on the distribution of the training and testing data, we divide the evaluation into three settings as shown in Table~\ref{table:dataset}: 
\textbf{\textit{In-distribution setting}} evaluates the model that resembles what it has seen during training. 
% Our training data comes from the MATH~\citep{hendrycks2021measuring} dataset, and we report the results from its test set as the in-domain evaluation.
\textbf{\textit{Few-shot setting}} aims to evaluate whether our proposed method can effectively learn from a small amount of labeled data. In the real world, it is often difficult to obtain large amounts of in-domain training data across different tasks, but a small number of cases can be annotated. 
% Therefore, we selected 4 examples from the training set of each sub-category from BBH~\citep{suzgun2022bbhchallenging}, Game of 24~\citep{yao2024tree}, and TheoremQA~\citep{chen2023theoremqa} separately to train our model and test its effectiveness on the test set or hold-out set.
\textbf{\textit{Out-of-distribution (OOD) setting}} further evaluates whether the model can handle scenarios it was not explicitly trained for, testing its ability to generalize beyond the training set. 
% The OOD datasets include Deepmind Math~\citep{saxton2018deepmindanalysing}, MMLU-pro~\citep{wang2024mmlupro}, strategyQA~\citep{geva2021didsqa}, and DROP~\citep{dua2019drop}.
For the training data, we use the MATH~\citep{hendrycks2021measuring} training set. For the few-shot learning, we select 4 examples from each category of BBH~\citep{suzgun2022bbhchallenging} as it is composed of 27 diverse tasks,\footnote{\url{https://huggingface.co/datasets/lukaemon/bbh}} resulting in 108 examples in total, 4 examples from Game of 24~\citep{yao2024tree}, and 4 examples from TheoremQA~\citep{chen2023theoremqa} datasets.
% For the training data, we use the MATH~\citep{hendrycks2021measuring} training set. and for the few-shot learning, we select 4 examples from each sub-task of BBH~\citep{suzgun2022bbhchallenging}, Game of 24~\citep{yao2024tree}, and TheoremQA~\citep{chen2023theoremqa} datasets. 
For the test data, we evaluate the model on the test set of the MATH dataset for the in-distribution setting and on the test sets or hold-out sets of BBH, Game of 24, and TheoremQA for the few-shot learning setting. For the OOD evaluation, we test each approach’s generalization ability on Deepmind Math~\citep{saxton2018deepmindanalysing}, MMLU-pro~\citep{wang2024mmlupro}, strategyQA~\citep{geva2021didsqa}, and DROP~\citep{dua2019drop}.
% \zyc{All evaluations (unless specified) were conducted when prompting the solver LLMs in zero shot.}
All evaluations (unless specified) were conducted when prompting the solver LLMs in zero shot.
For answer evaluation, we use the simple-eval\footnote{\url{https://github.com/openai/simple-evals}.} for MATH, a standard evaluation for Game of 24~\citep{yao2024tree}, and exact string matching for the others.
% \zyc{todo: is deepmind math really OOD? Also be clear: in-distribution rather than in-domain; out of distribution rather than out of domain.}

\vspace{-1em}
\paragraph{Training Setup}
% In our optimal trajectory search, we set the number of iterations ($K$) to 2 and the number of evaluation times ($N_{eval}$) to 4. The number of paths retained $N_{1}$ is set to 8 and $N_{2}$ to 3. Throughout the search, we maintain a sampling temperature of 0.4. 
For external planner tuning, we utilize Llama-3-8B-Instruct as our planner and GPT-4o-mini and Llama-70B-Instruct as task-solving LLMs. 
Experiments of internalized planner tuning were conducted with Llama-3-8B-Instruct. For more details, refer to Appendix~\ref{Appendix: implementation}.

\vspace{-1em}
\subsection{Baselines}
\vspace{-0.5em}
We include the following highly related baselines in our experiments.
(1) CoT~\citep{wei2022chain} prompts an LLM to answer step-by-step; (2) PoT~\citep{chen2022program} prompts an LLM to generate Python code and execute the code to get the final answer; (3) Least-to-most (LTM)~\citep{zhou2023leasttomost} prompts an LLM to first decompose the question into multiple sub-questions before solving it; 
% (4) Self-refine~\citep{madaan2024self} prompts an LLM to generate the answer with code and verify the answer by the LLM itself, then refine the answer; 
(4) Self-refine~\citep{madaan2024self} prompts an LLM to generate the answer and verify and refine the answer by the LLM itself. \cite{madaan2024self} used PoT in solving math questions, therefore we follow their setting to use PoT in generating the initial answer; 
% (5) Automatic Prompting Optimization (APO)~\citep{wang2023promptagent} searches for a better prompt for the specific task based on its training data; this baseline is fine-tuned on the same sources of training data as our approach with its default hyperparameter setting.
(5) PromptAgent (PA)~\citep{wang2023promptagent} searches for a better prompt for the specific task based on its training data;
this baseline is implemented with the default hyperparameter setting;
% this baseline is fine-tuned with its default hyperparameter setting;
% this baseline is fine-tuned on the same sources of training data as our approach with its default hyperparameter setting.
% \zyc{"APO" to "PromptAgent". check my revision; you may briefly clarify how the training set was used to fine-tune PromptAgent.} 
% In our experiment, we leverage the same training data for our method and the default hyperparameters. 
and (6) Vanilla Supervised Fine-Tuning
(Vanilla SFT) uses GPT-4o to generate the CoT reasoning process for questions in the training datasets and then fine-tune the solver LLM to predict the generated reasoning process and the ground-truth answer; this baseline is fine-tuned using the same hyperparameter setting as our internalized planner tuning. 
% We follow the same training implementation e.g., hyperparameters, in Vanilla SFT as ours. 
The training data for PA, Vanilla SFT, and \method are from the same source.  
% \zyc{todo: add a math formula for Vanilla SFT. Unclear what "the same training implementation" means. Need to clarify that the examples came from the same \emph{source} of training data.}\murong{changed the statement}

\vspace{-1em}
\subsection{External Planner Tuning Results}
\vspace{-0.5em}
\begin{table}[t!]
\vspace{-1em}
\caption{Accuracy (\%) of the external planner tuning on in-distribution and few-shot datasets. The reasoning format $\mathcal{L}$ represents language, and $\mathcal{P}$ means program. 
%The highest scores for each column are in bold format.
}\vspace{2mm}
\centering
\small
\setlength{\tabcolsep}{3pt}
\begin{tabular}{lcccccc||c}
\hline
\multirow{2}{*}{Method} & \multirow{2}{*}{Tuning} & Reasoning  & \multirow{2}{*}{MATH} & \multirow{2}{*}{BBH} & \multirow{2}{*}{Game of 24} & \multirow{2}{*}{TheoremQA} & \multirow{2}{*}{Average}  \\
 & & Format & && & & \\
\hline\hline
\multicolumn{8}{l}{\textbf{External Planner: Llama-3-8B-Instruct; Solver: Llama-3-70B-Instruct}} \\
\hline

CoT &\ding{55} & $\mathcal{L}$ & 50.4 & 72.7 & 27.5 & 27.4 & 44.5\\
LTM &\ding{55} &$\mathcal{L}$ & 50.1 & 73.8 & 24.9  
& 28.8 & 44.4 \\
PA &\ding{51} &$\mathcal{L}$& 52.5 & 72.9 & 26.8 & 28.8 & 45.3 \\
PoT &\ding{55} &$\mathcal{P}$ & 54.7 & 65.8 & 63.9 & 31.1 & 53.9 \\
Self-refine &\ding{55} &$\mathcal{L}$, $\mathcal{P}$ & 55.9 & 71.4 & \bf{68.3} & 30.8 & 56.6 \\
\textbf{\method: External} &\ding{51} &$\mathcal{L}$, $\mathcal{P}$ & \bf{57.7} & \bf{77.3} & 67.7 & \bf{31.2} & \bf{58.5} \\
\hline\hline
\multicolumn{8}{l}{\textbf{External Planner: Llama-3-8B-Instruct; Solver: GPT4o-mini}} \\
\hline
CoT &\ding{55} &$\mathcal{L}$ & 70.2 & 80.3 & 27.7 & 38.9 & 54.2 \\
LTM &\ding{55} &$\mathcal{L}$ & 72.2 & 79.4 & 25.5 & 36.4 & 53.3 \\
PA &\ding{51} &$\mathcal{L}$& 73.5 & 81.1 & 26.7& 38.9 & 55.1 \\
PoT &\ding{55} &$\mathcal{P}$ & 67.2 & 73.9 & 61.4 & 35.8 & 59.6 \\
Self-refine &\ding{55} &$\mathcal{L}$, $\mathcal{P}$ & 73.7 & 74.8 & \bf{68.7} & 34.6 & 63.0 \\
\textbf{\method: External} &\ding{51} &$\mathcal{L}$, $\mathcal{P}$ & \bf{75.4} & \bf{84.2} & 65.2 & \textbf{41.4} & \bf{66.5} \\
\hline
\end{tabular}
\vspace{-1em}

\label{table:in-domain external}
\end{table}
Table~\ref{table:in-domain external} presents the results of using the external planner, which suggest that:
\vspace{-1em}
\paragraph{External planner tuning outperforms other methods on the in-domain task} 
Our method achieves 57.7\% accuracy with Llama-3-70b-Instruct and 75.4\% accuracy with GPT-4o-mini on MATH, achieving significant improvement than baselines. 
This suggests that \method is robust across different LLMs and it can significantly enhance the LLM's zero-shot reasoning ability.
The improvement from \method remains consistent as the solver LLM's capabilities increase, indicating \method has a long-term value even as LLMs continue to improve rapidly.
% From the first column of Table~\ref{table:in-domain external}, static prompt engineering methods offer limited improvement. In contrast, our method achieves 57.7\% accuracy with Llama-3-70b-Instruct and 75.4\% accuracy with GPT-4o-mini on MATH by enabling the LLM to dynamically select actions based on the question and capability of the task-solving LLM. 
% Notably, the performance improvements from our \method approach are greater with GPT-4o-mini compared to Llama-3-70B-Instruct, likely due to GPT4o-mini's superior inherent reasoning capabilities which allow it to leverage the optimal combination of reasoning actions more effectively.
% \wenlin{I looked at Table 3 again and it is interesting that GPT-4o-mini improves more using our method than Llama-3-70B, probably due to its superior inner reasoning capabilities. Maybe we can discuss more here?}
% \wenlin{Notably, the performance improvements from our DOTS approach are greater with GPT4o-mini compared to Llama-3-70B-Instruct, likely due to GPT4o-mini's superior inherent reasoning capabilities which allow it to leverage the optimal combination of reasoning actions more effectively.}
\vspace{-1em}
\paragraph{The external planner can learn the appropriate action trajectory with only a few training examples.} On the BBH, \method achieves improvements of 3.5\% and 3.1\% over the best static methods when using Llama-3-70B-Instruct and GPT-4o-mini, respectively. In the Game of 24 and TheoremQA, \method also shows slight improvements or performs similarly to the best static method. 
This indicates that even a small number of cases can help the LLM learn the optimal strategy for the given task.
Besides, \method demonstrates greater stability across various datasets.
% Besides, \method demonstrates greater stability, either exceeding or closely matching the best method across various datasets.
Our flexible action trajectory selection demonstrates its advantages on datasets requiring diverse reasoning actions, such as BBH as shown in Appendix~\ref{Appendix: bbh}.
Conversely, the Game of 24 features a uniform question type, where the predefined static method self-refine is sufficient.
While the self-refine excels on Game of 24, it significantly lags behind on other datasets.
This reflects the external planner’s ability to effectively select the appropriate action trajectory, leading to more robust performance even across tasks with varying reasoning demands.

\vspace{-1em}
\subsection{Internalized Planner Tuning Results}
\vspace{-0.5em}
Table~\ref{table: in-domain internal} presents the results of our internalized planner tuning, where we observed:
\vspace{-1em}
\paragraph{Internalized planner tuning demonstrates superior performance} 
\method outperforms existing methods on average, including prompt engineering methods and vanilla SFT. 
% Our method demonstrated superior overall performance, outperforming existing methods on average, including prompt engineering methods and vanilla SFT. 
Notably, our approach surpasses self-refine in the Game of 24, a different observation than the experiments with an external planner (Table~\ref{table:in-domain external}).
We attribute this performance boost to our joint optimization of the trajectory planning and problem-solving processes. Unlike external planner tuning which only updates the external planner ($LLM_p$), internalized planner tuning enables the task-solving LLM to simultaneously learn trajectory planning and accurate reasoning process generation. 
This highlights that the internalized planner tuning effectively further enhances performance.
\vspace{-1em}
\paragraph{Searching for the optimal reasoning action trajectory helps enhance the utilization of training data} Compared to vanilla SFT, our method consistently shows performance improvements across all datasets, notably achieving an 8.7\% increase on BBH. This suggests that, instead of training with a question and step-by-step reasoning process pair, our approach of searching for an optimal action trajectory and generating the corresponding reasoning process to construct training data is superior. 
This finding indicates that our search methodology could effectively enhance the utilization of training data for reasoning tasks without the need for additional human annotations.
\begin{table}[htbp]
\vspace{-1em}
\caption{Internal planner tuning performance on in-distribution and few-shot datasets.}\vspace{2mm}
\centering
\small
\setlength{\tabcolsep}{3pt}
\begin{tabular}{lcccccc||c}
\hline
% \multirow{2}{*}{Method} & \multirow{2}{*}{Tuning}  & Reasoning  & MATH & BBH & Game of 24 & TheoremQA & \multirow{2}{*}{Average} \\
\multirow{2}{*}{Method} & \multirow{2}{*}{Tuning} & Reasoning  & \multirow{2}{*}{MATH} & \multirow{2}{*}{BBH} & \multirow{2}{*}{Game of 24} & \multirow{2}{*}{TheoremQA} & \multirow{2}{*}{Average}  \\
 & & format & & &  &  & \\
\hline\hline
\multicolumn{8}{l}{\textbf{Solver: Llama-3-8B-Instruct}} \\
\hline
CoT &\ding{55} &$\mathcal{L}$ & 29.6 & 48.9 & 12.7 & 14.8 & 26.5 \\
LTM &\ding{55} &$\mathcal{L}$ & 29.5 & 50.3 & 14.4 & 15.2 & 27.4 \\
PA &\ding{51} &$\mathcal{L}$ & 31.0 & 47.2 & 11.8 & 15.1 & 26.3 \\
PoT &\ding{55} &$\mathcal{P}$ & 25.3 & 44.6 & 16.8 & \bf{16.7} & 25.9 \\
Self-refine &\ding{55} &$\mathcal{L}$, $\mathcal{P}$ & 28.7 & 46.6 & 17.0 & 15.3 & 30.1 \\
Vanilla SFT &\ding{51} &$\mathcal{L}$ & 33.9 & 61.0 & 18.5 & 14.8 & 33.6 \\
\textbf{\method: Internalized} &\ding{51} &$\mathcal{L}$, $\mathcal{P}$ & \bf{34.4} & \bf{69.7} & \bf{21.9} & 16.1 & \bf{35.5} \\
\hline
\end{tabular}
\vspace{-1em}

%: MATH, BBH, Game24, and TheoremQA.}
\label{table: in-domain internal}
\end{table}

\vspace{-1em}
\subsection{Out-of-Distribution Experimental Results}
\begin{table}[t!]
\vspace{-1em}
\caption{
%Performance 
Accuracy (\%)
on out-of-distribution (OOD) tasks.}
\vspace{3mm}
\centering
\small
\setlength{\tabcolsep}{3pt}
\begin{tabular}{lcccc||c}
\hline
Method & DeepMind-Math & MMLU-pro & StrategyQA & DROP & Average \\
\hline
\hline
\multicolumn{6}{l}{\textbf{External Planner: Finetuned Llama-3-8B-Instruct; Solver: Llama-3-70B-Instruct}} \\
\hline
CoT  & 54.6 & 60.6 & 81.3 &  66.1 & 65.6 \\
LTM & 55.6 & \bf{60.9} & \bf{81.9} &64.3& 65.6 \\
PA & 58.1 & 54.2 & 80.3 & 58.7&62.8 \\
PoT  & 73.0 & 57.3 & 74.8 & 62.8 & 66.9 \\
Self-refine  & 73.9 & 59.5 & 77.8 &64.8 &69.0 \\
\textbf{\method: External} & \bf{74.1} & 59.4 & 80.3 & \bf{66.3} & \bf{70.0} \\
\hline
\hline
\multicolumn{6}{l}{\textbf{External Planner: Finetuned Llama-3-8B-Instruct; Solver: GPT4o-mini}} \\
\hline
CoT & 80.2 & \bf{61.7} & 78.8 &  65.8 & 71.6 \\
LTM & 80.6 & 61.4 & \bf{80.9} & 64.5& 71.8\\
PA & 82.2 & 48.1 & 78.3 & 67.0& 68.9 \\
PoT  & \bf{87.7} & 57.1 & 77.9 &  72.4 & 73.7 \\
Self-refine  & 85.9 & 58.3 & 77.2 & 72.3 & 73.4\\
\textbf{\method: External} & 87.6 & 61.5 & 78.8 & \bf{73.8} & \bf{75.4} \\
\hline
\hline
\multicolumn{6}{l}{\textbf{Solver: Finetuned Llama-3-8B-Instruct}} \\
\hline
CoT & 28.3 & 37.2 & \bf{72.7} &  52.9 & 47.8 \\
LTM & 30.9 & 38.6 & 70.7 &  \bf{55.2} & 48.9 \\
PA & 29.3 & 34.5 & 69.7 &  51.6 & 46.3 \\
PoT  & 48.1 & 37.3 & 63.9 & 44.6&48.5 \\
Self-refine  & 44.9 & 33.1 & 65.3 & 47.1&47.6 \\
Vanilla SFT & 39.6 & \bf{40.3} & 71.8 & 49.0 &50.2 \\
\textbf{\method: Internalized} & \bf{55.3} & 39.7 & 68.2 & 48.8 & \bf{53.0} \\
\hline
\end{tabular}
\vspace{-1em}

\label{table:ood-performance}
\end{table}
% \paragraph{Enabling LLM to actively think the best action trajectory improves reasoning results in OOD tasks} 
% \paragraph{\method outperforms the baseline models and at least does not hurt the performance for unseen tasks}\zyc{what does this mean -- "at least does not hurt the performance"? we do have lower results than baselines.}
% From Table~\ref{table:ood-performance}, we learn that our method maintains a good generalization ability across OOD tasks. Specifically, for the external planner tuning, our method achieved or approached the best performance across all test datasets. For internalized planner tuning, our tuned model outperformed the baseline models on the DeepMind Math and MMLU-pro datasets. However, its performance was slightly lower than the baseline models on other datasets. This performance difference is primarily due to the significant misalignment between the types of tasks in these datasets (common sense reasoning, etc.) and the type in our training data (math reasoning).
\vspace{-0.5em}
\paragraph{Our method consistently generalizes well across diverse OOD challenges} As shown in Table~\ref{table:ood-performance}, \method maintains high accuracy across different datasets and models. In contrast, static methods often fluctuate significantly in performance. For instance, despite static methods like CoT showing a slight advantage on MMLU-Pro and StrategyQA over \method using the Llama-3-70B-Instruct model, they experience a sharp decline on DeepMind Math. 
% Specifically, on the DeepMind-Math dataset, LTM performs poorly (55.6\% vs \method 74.1\%), reflecting 18.5\% lower than our method. 
This pattern of fluctuations can be observed in other methods as well, where some excel on individual tasks but fail to maintain strong performance. 
In contrast, \method continues to deliver consistently high accuracy across various models and datasets. The stability of our method is attributed to its ability to dynamically select appropriate reasoning trajectories. The results indicate that \method is better suited to meet the demands of diverse tasks, demonstrating stronger robustness and generalization, making it a more reliable and adaptable approach for handling a wide variety of OOD challenges.

\vspace{-1em}
\subsection{Ablation Study}
\begin{table}[htp]
\vspace{-2mm}
\caption{Ablation Study}
\vspace{2mm}
\centering
\small
\begin{tabular}{lcccc||c}
\hline
                    & MATH & BBH  & Game24 & TheoremQA & Average \\ \hline
\multicolumn{6}{l}{\bf External Planner: Llama-3-8B-Instruct; Solver: GPT-4o-mini} \\ \hline
\method: External          & 75.4 & 84.2 & 65.2   & 42.4 & 66.8   \\  
-w/o Searching     &  69.2    &  78.6      &   28.9   &  40.2  &   54.2   \\
-w/o Explanation     &   68.2   &  81.3    &   57.4     &  36.4    &    60.8     \\
\hline
\multicolumn{6}{l}{\bf Internalized Planner \& Solver: Llama-3-8B-Instruct} \\ \hline
\method: Internalized          & 34.4 & 69.7 & 21.9   & 16.1 & 35.5   \\ 
-w/o Searching &  31.4    &  55.8    &   19.6     &  15.1    & 30.5        \\
-w/o Explanation &  33.8  &  65.8    &  18.6    &    15.7    &   33.4         \\
\hline
\end{tabular}
\vspace{-1em}

\label{table: ablation}
\end{table}

In this section, we perform the ablation study and assess the effectiveness of each component of our method: (1) \textbf{Without Searching:} To demonstrate the effectiveness of searching for the optimal action trajectory, we test the performance of the LLM tuned with a randomly selected action trajectory; (2) \textbf{Without Explanation:} To understand if training the planner to generate an explanation for the optimal reasoning trajectory is helpful, we test \method's performance when the planner is trained to predict the trajectory without explanation.
% To evaluate the impact of explanation generation, we tested the performance when, after identifying the optimal action trajectory, no explanation was generated. The LLM was directly fine-tuned on the optimal trajectory without any accompanying explanations. 

The results in Table~\ref{table: ablation} indicate that both optimal trajectory searching and explanation generation are crucial in \method. 
% The first part of the table reveals that when the model is presented with random action sequences, the LLM fails to learn the selection of the best action trajectory.
% From the first part of the table, we find that the LLM's ability to learn or generalize is compromised when presented with random action sequences. 
For example, in the Game of 24, the planner trained without searching for the optimal trajectory did not consistently select the PoT action (which was considered the most effective for this task) in its trajectory. 
% This highlights the necessity of performing a search, as it shows that the LLM learns the relationship between tasks and the optimal action trajectory, rather than merely adapting to a specific answering format.
% This highlights the necessity of performing a search and shows that the LLM has learned the relationship between tasks and best action trajectory, rather than just adopting to follow the answering format. 
Additionally, we observe that without explanations, the planner's ability to predict optimal trajectories becomes less reliable. Incorporating explanations effectively guides the planner to learn to predict suitable action trajectories for the given questions.
% For explanations, we find that the model's ability to predict action trajectories became less reliable. Adding explanations effectively guides the LLM to better predict appropriate action trajectories for given tasks.
\vspace{-1em}
\subsection{Optimal Trajectory Analysis for Different Tasks}
% \input{tables/distribution_v3}
% Comments (haitaomi): let's add one more column at the very beginning: score on each sub-task, and sorted from easy to hard, then we can have a better idea how planing pattern changes as task changes from easy to hard.
\vspace{-0.5em}
Table~\ref{table: action distribution} shows the distribution of actions selected in the optimal trajectories by our planner on the MATH test set. The distribution suggests two key findings:
\vspace{-1em}
\paragraph{\method adapts to the characteristics of specific questions} In mathematics, number theory problems are more suitable to be solved with programs, so the proportion of PoT is higher, while geometry problems are not easily represented and solved with naive Python code; as a result, our planner mainly uses CoT for such problems. This indicates that \method tailors its action selection based on the unique characteristics of each problem type.
\vspace{-1em}
\paragraph{\method adapts to the capability of specific task-solving LLMs} As shown in Table~\ref{table:in-domain external}, on the MATH dataset, GPT-4o-mini performs better using CoT for problem-solving, whereas Llama3-70B-instruct performs better using PoT. 
When GPT-4o-mini is the task-solving LLM, our fine-tuned planner selects a higher proportion of CoT actions; when Llama3-70B-Instruct is used, PoT actions dominate. This suggests that our planner is not only aware of the problem type but also adapts the reasoning action trajectory prediction based on the capabilities of the task-solving LLM.
% From the distribution, we can see that when GPT-4o-mini is the task-solving LLM, our fine-tuned planner selects a higher proportion of CoT, whereas when Llama3-70b-Instruct is the task-solving LLM, PoT accounts for a higher proportion. It indicates that the selection of reasoning actions by our planner considers the inherent capability of the task-solving LLM.

Furthermore, we observe that question rewriting and decomposition were selected with a low frequency. 
This is likely because the MATH dataset consists of precise problems that do not benefit from rewriting.
Additionally, given the strong reasoning abilities of Llama3-70B-Instruct and GPT-4o-mini, their CoT process inherently includes task decomposition, reducing the need for further planning interventions.
% \usepackage{multirow}
\begin{table}[]
\vspace{-1em}
\caption{Planning action distributions of \method over three different layers on the MATH test set. 
% Rewr. is the abbreviation of Rewriting, Deco. means Decomposition, and Veri. means Verification.
}
\vspace{2mm}
\centering
\small
\setlength{\tabcolsep}{3pt}
\resizebox{0.8\textwidth}{!}{
\begin{tabular}{l|c|ccc|cc|cc}
\hline
\multirow{2}{*}{Sub-tasks on MATH} & \multirow{2}{*}{Accuracy (\%)} & \multicolumn{3}{c|}{Analysis Layer} & \multicolumn{2}{c|}{Solution} & \multicolumn{2}{c}{Verification} \\ 
\cline{3-9} 
                             &                         &  Rewr. & Deco. & Empty              & CoT & PoT                        & Veri. & Empty \\ 
\hline \hline
\multicolumn{9}{l}{\textbf{External Planner: Llama-3-8B-Instruct; Solver: GPT-4o-mini}} \\ 
\hline
Algebra      & 92.1 & 0.03 & 0.05 & 0.92 & 0.90 & 0.10 & 0.29 & 0.71 \\
Prealgebra   & 88.6 & 0.03 & 0.01 & 0.96 & 0.79 & 0.31 & 0.21 & 0.79 \\
  {Number Theory} &
  81.8 &
  0.01 &
  0.01 &
  {0.98} &
  0.43 &
  {0.57} &
  0.15 &
  0.85 \\
  {Counting and Probability} &
  76.8 &
  0.08 &
  0.06 &
  {0.84} &
  0.78 &
  {0.32} &
  0.30 &
  0.70 \\
  {Geometry} &
  61.8 &
  0.03 &
  0.01 &
  {0.96} &
  0.95 &
  {0.05} &
  0.06 &
  0.94 \\
  {Intermediate Algebra} &
  57.1 &
  0.05 &
  0.02 &
  {0.93} &
  0.85 &
  {0.15} &
  0.44 &
  0.56 \\
  {Precalculus} &
  52.6 &
  0.06 &
  0.02 &
  {0.92} &
  0.95 &
  {0.05} &
  0.46 &
  0.54 \\ \hline \hline
\multicolumn{9}{l}{\textbf{External Planner: Llama-3-8B-Instruct; Solver: Llama-3-70B-Instruct}} \\ \hline 
  {Algebra} &
  {74.9} &
  0.03 &
  0.04 &
  {0.93} &
  0.77 &
  {0.23} &
  0.12 &
  0.88 \\
  {Prealgebra} &
  {74.5} &
  0.02 &
  0.03 &
  {0.95} &
  0.57 &
  {0.43} &
  0.10 &
  0.90 \\
  {Number Theory} &
  {69.9} &
  0.01 &
  0.01 &
  {0.98} &
  0.32 &
  {0.68} &
  0.13 &
  0.87 \\
  {Counting and Probability} &
  {55.4} &
  0.04 &
  0.02 &
  {0.94} &
  0.59 &
  {0.41} &
  0.11 &
  0.89 \\
  {Geometry} &
  {39.6} &
  0.05 &
  0.01 &
  {0.94} &
  0.76 &
  {0.24} &
  0.18 &
  0.82 \\
  {Precalculus} &
  {36.9} &
  0.07 &
  0.03 &
  {0.90} &
  0.78 &
  {0.22} &
  0.28 &
  0.76 \\
  {Intermediate Algebra} &
  {34.6} &
  0.03 &
  0.01 &
  {0.96} &
  0.72 &
  {0.28} &
  0.20 &
  0.80 \\ \hline
\end{tabular}
}
\vspace{-1em}

\label{table: action distribution}
\end{table}

% \zyc{interesting. do we have statistics for trajectories collected from MATH? Would a training set better cover all cases when each action is helpful?}
\vspace{-1em}
\subsection{Additional Analyses}
\vspace{-0.5em}
\paragraph{Few-shot In-context Learning Setting} 
\begin{wraptable}{r}{0.55\textwidth}
\centering
\small
\vspace{-3em}
\caption{External planner tuning under the few-shot setting with GPT-4o-mini as the solver.}
\vspace{1em}
\label{table: few-shot}
\begin{tabular}{lcccc}
\hline
Method        & MATH          & BBH           & TheoremQA     & Average       \\ \hline
CoT           & 72.3          & 84.2          & 38.2          & 64.9          \\
LTM           & 72.7          & 83.4          & 37.3          & 64.5          \\
PA           & 71.3          & 83.3          & 38.7 & 64.4          \\
PoT           & 69.8          & 82.1          & 36.4          & 62.8          \\
Self-refine   & 73.2          & 83.1          & 35.4          & 63.9          \\
\textbf{\method} & \textbf{75.4} & \textbf{86.1} & \textbf{39.9}          & \textbf{67.1} \\ \hline
\end{tabular}
\vspace{-1em}
\end{wraptable}
Our main results report the performance with zero-shot evaluation. In cases where reasoning tasks are known in advance, a common approach to leveraging training data and improving the performance of closed-source LLMs is few-shot in-context learning (ICL), where training examples are incorporated directly into the context. 
Our external planner tuning can also be utilized in this scenario seamlessly. 
Specifically, we can first construct few-shot ICL prompts for each potential reasoning action trajectory. Once the external planner selects the appropriate reasoning actions, the corresponding few-shot prompt will be chosen and applied.
% We could first construct few-shot ICL prompts for every possible reasoning action trajectory. 
% Once the actions are determined by the external planner, the few-shot prompt can be selected accordingly.
% In this section, we evaluate the performance of our external planner method in the few-shot setting using a fine-tuned Llama-3-8B-Instruct as the external planner and GPT-4o-mini as the solver LLM and compare it with other approaches. We randomly selected 8 examples from MATH, 4 examples from each subcategory of the BBH, and 4 examples from the TheoremQA dataset.\footnote{We excluded the "Game of 24" task because knowing the task in advance allows it to be solved with a simple program.} All few-shot demonstrations were generated by GPT-4o and manually verified for quality.
% In this section, we evaluate the performance of our external planner method in the few-shot setting, using a fine-tuned Llama-3-8B-Instruct as the external planner and GPT-4o-mini as the solver LLM. 
We evaluate the external planner tuning setup of \method, with Llama-3-8B-Instruct being the external planner and GPT-4o-mini being the solver LLM, in this setting.
We compare our approach with the same baselines similarly implemented in the few-shot ICL setting, where we randomly selected 
% other methods by randomly selecting 
8 examples from MATH, 4 examples from each category of BBH, and 4 examples from TheoremQA to form the prompt.\footnote{We excluded the ``Game of 24'' task because knowing the task in advance enables it to be solved with a straightforward program.} All few-shot demonstrations were generated by GPT-4o and manually verified for quality.

As shown in the Table~\ref{table: few-shot}, \method continues to outperform baseline models. Interestingly, compared to Table~\ref{table:in-domain external}, which presents the zero-shot results, adding few-shot demonstrations to static prompting methods does not lead to consistent improvement, except on the BBH dataset. This indicates that simply expanding the context with additional demonstrations does not always serve as an effective way to leverage available training data. 
In contrast, our method demonstrates its superior ability to effectively utilize the training data.

% As shown in the table~\ref{table: few-shot}, \method continues to outperform baseline models. Interestingly, only the performance on the BBH dataset improved after adding a few shot examples. This indicates that simply adding training data to the context does not always benefit current powerful LLMs and complex tasks, but \method consistently improves performance, demonstrating that it can better utilize training data.

% \vspace{-1em}
% \paragraph{Case study} 
% We conducted a case study to provide intuitive examples to demonstrate the effectiveness of our method. As shown in Appendix~\ref{Appendix: case}, we can see that our model is capable of actively selecting appropriate actions for reasoning problems.

\vspace{-1em}
\newpage
\paragraph{How efficient is \method?}
\begin{wraptable}{r}{0.45\textwidth}
\vspace{-1em}
\caption{Avg. number of output tokens for each method (solver: Llama-3-8B-Instruct).}
\vspace{2mm}
\centering
\small
\setlength{\tabcolsep}{3pt}
\begin{tabular}{l|c}
\hline
\multirow{2}{*}{Method}      & Avg. \# of \\
                            & Output Tokens \\ \hline
CoT~\citep{wei2022chain}         & 263.6               \\ \hline
LTM~\citep{zhou2023leasttomost}         & 436.4                \\ \hline
Self-refine~\citep{madaan2024self}  & 527.6              \\ \hline
\textbf{\method: Internalized} & 409.1                \\ \hline
\end{tabular}
% \vspace{-2em}

\label{table:token}
\end{wraptable}

% CoT~\citep{wei2022chain} &\ding{55} &\ding{55} &\ding{51} &\ding{55} &\ding{55} \\
% PoT~\citep{chen2022program} &\ding{55} &\ding{55} &\ding{55} & \ding{51}&\ding{55} \\
% LTM~\citep{zhou2023leasttomost} &\ding{51} &\ding{55} &\ding{51} &\ding{55} &\ding{55} \\
% R\&R~\citep{deng2023rephrase} &\ding{55} & \ding{51}&\ding{51} &\ding{55} &\ding{55} \\
% Self-Refine~\citep{madaan2024self} &\ding{55} &\ding{55} &\ding{55} &\ding{51} & \ding{51}\\
% We compared \method's cost efficiency of the fine-tuned Llama-3-8b-Instruct after internalized planner tuning to other prompt engineering techniques using the Llama-3-8b-Instruct without tuning. This was measured as the average output token count. 
We compare the cost efficiency, measured by the average output token count, of each method (based on Llama-3-8B-Instruct) in Table~\ref{table:token}.
The result shows that \method consumes fewer tokens on average than other advanced approaches and only more than CoT.
% \zyc{is it fair to say "slightly"?} 
% This is because our method can choose different actions for each problem, reducing tokens.
Advanced prompt engineering methods often introduce supplementary text to facilitate reasoning. However, not all questions require this additional context to the same extent. By constructing training data via searching, our goal is to optimize the balance between minimizing extraneous steps and maintaining a high success rate, thereby reducing unnecessary output tokens. Our method avoids redundant reasoning actions, resulting in a more efficient system.

% \subsection{Case Study}
% % \input{tables/case}
% In this section, we conducted a case study to provide intuitive examples to demonstrate the effectiveness of our method. As shown in Appendix~\ref{Appendix: case}, we can see that our model is capable of actively selecting appropriate actions for reasoning problems. 
% For more cases, please refer to the Appendix~\ref{Appendix: case}.
% $Q_{II}$ is a question widely tested in the community where GPT-4o even tends to make mistakes, but after tuning with searched trajectory data, the LLM actively chose to use code to solve the problem.

% \subsection{How efficient is our method?}
% \input{tables/output_token}
% % Comments: make this paragraph short and simple. 
% In this section, we validate that our method is more efficient compared to other prompt engineering methods. We use the average output token count as the metric for efficiency. For the same base model, LLaMa-3-8b-instruct, we compared our method with other prompt engineering approaches on the test dataset in terms of token number. The results in Table~\ref{table:token} show that our average token count is better than other prompt engineering techniques. This is primarily because our method can dynamically select actions, avoiding the token waste caused by using the same action for all problems.
\vspace{-1em}
\paragraph{Do we need more reasoning steps for difficult questions?}
\begin{wrapfigure}{R}{0.45\textwidth}
\vspace{-1em}
\caption{Average reasoning trajectory length per difficulty level on MATH for \method (solver: GPT-4o-mini; External planner: Llama3-8B-Instruct).
\vspace{3mm}
    \centering
    \includegraphics[width=0.35\textwidth]{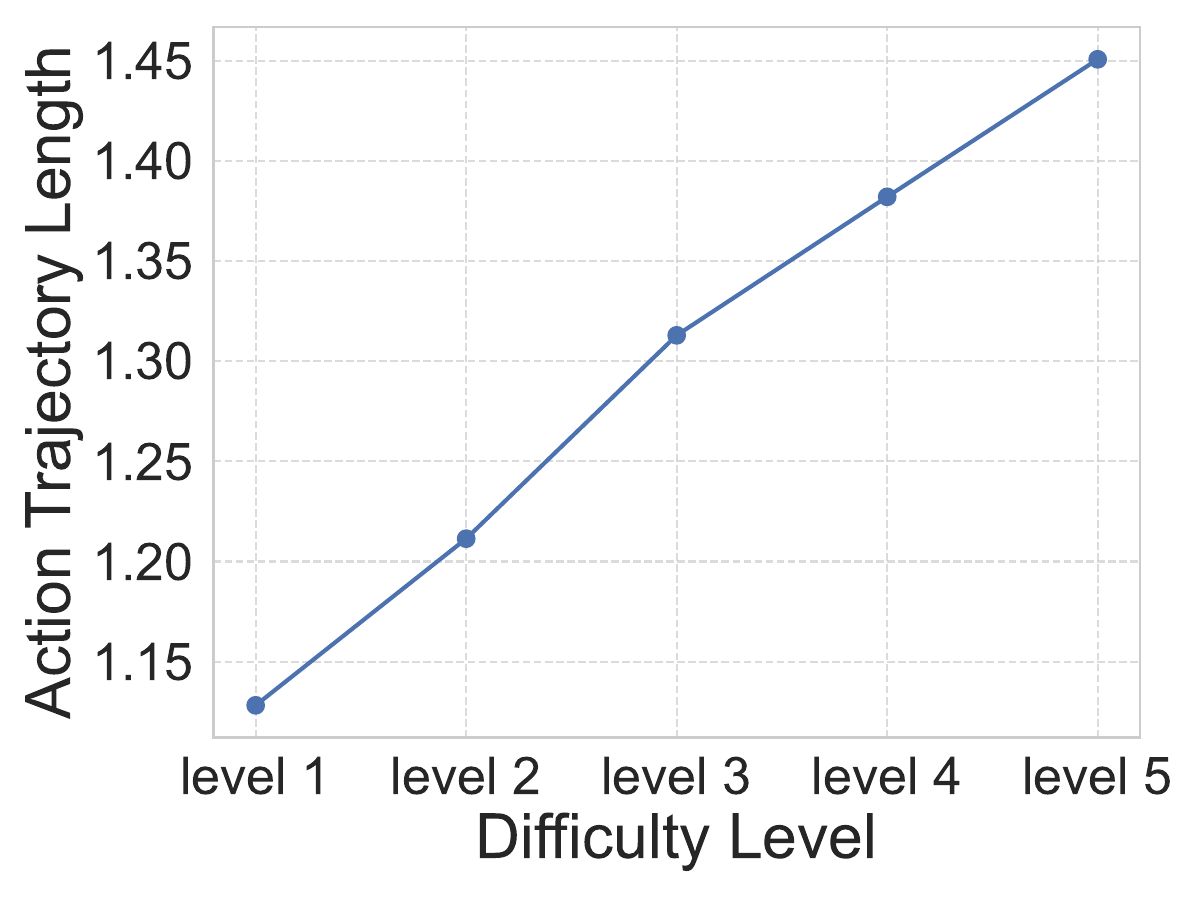}
    }
    \label{fig:difficulty}
    % \vspace{-4mm}
\vspace{-2em}

\end{wrapfigure}
Recent research suggests that LLMs can better solve difficult questions by increasing the thinking time in the inference stage~\citep{brown2024largemonkey,openai2024learning}.
% \zyc{you may want to create a bib item for the footnote link} 
In our study, we explore the relationship between question difficulty and the average reasoning action trajectory length. 
The trajectory length is determined by assigning a value of $0$ to the \texttt{EMPTY} module and $1$ to all other actions, while the question difficulty is derived from annotated levels on the MATH dataset.
% We employ GPT-4o-mini as the task-solving LLM and the fine-tuned Llama3-8b-Instruct as the external planner. 
Figure~\ref{fig:difficulty} presents that harder problems demand more computational steps, resulting in longer reasoning trajectories. Case analyses further reveal that our planner increases the proportion of verification steps as problem difficulty rises. This highlights an exciting fact --- LLMs can learn to employ more reasoning steps for challenging problems through exploration, without requiring explicit expert guidance.
\vspace{-1em}
\section{Related Work}
\vspace{-1em}
\paragraph{Prompt engineering for LLM reasoning} 
LLMs have demonstrated remarkable proficiency in solving complex reasoning tasks~\citep{rae2021scaling, lewkowycz2022solving, zhong2023agieval}. The Chain-of-Thought (CoT) approach, introduced by \citet{wei2022chain}, significantly improves performance on reasoning problems by prompting LLMs to think step-by-step, thereby activating their inherent reasoning capabilities~\citep{madaan2022text}. To further enhance LLMs' capabilities in mathematical and symbolic reasoning, \citet{chen2022program} and \citet{gao2023pal} proposed the Program-of-Thought prompting method, where code is used as an intermediate reasoning step. 
Advanced prompt engineering methods, such as question decomposition~\citep{zhou2023leasttomost} and self-verification~\citep{madaan2024self}, have also proven effective in improving reasoning performance. Additionally, recent approaches have incorporated automatic prompt optimization based on training data. For instance, \cite{wang2023promptagent} refines prompts by analyzing error cases, and self-discovery~\citep{zhou2024selfdiscovery} utilizes modular reasoning components to construct the task-adaptive prompt.
However, these automated prompt optimization techniques still produce static prompts for all instances.
Recently, \citet{srivastava-etal-2024-instances} proposed the instance-level prompt optimization via LLM self-refining while it is still a passive expert-designed workflow and lacks the explorations and evaluations to guide the LLM to better actively adapt to the question and LLM capability.
In our method, we internalize the reasoning action selection capability into the LLM itself without an expert-designed workflow, allowing it to autonomously fit both the characteristics of questions and the inherent capability of task-solving LLM.

\vspace{-1em}
\paragraph{Searching for boosting LLM reasoning} 
Recent research suggests that incorporating searching mechanisms can significantly enhance LLM reasoning. In the inference process, Tree-of-Thought (ToT)~\citep{yao2024tree} and Graph-of-Thought (GoT)~\citep{besta2024graph} have been proposed to search and investigate different reasoning paths, either by leveraging the LLM itself~\citep{yao2024tree} or designing heuristic functions~\citep{hao-etal-2023-reasoning} as the signal to evaluate each step. More recently, Monte Carlo Tree Search (MCTS) has been introduced to assist the LLM in learning how to evaluate each step~\citep{qi2024mutual,xie2024monte}. The searching mechanism can also be used in training to collect training instances for improving LLM reasoning~\citep{luo2024improve}. However, all these searching methods treat each ``CoT reasoning step'' as the atomic component or step in searching, while we choose each reasoning action as the atomic component in our case. 

\vspace{-1em}
\section{Conclusion}
\vspace{-1em}
In this paper, we introduce \method, a method that enables LLMs to autonomously think about appropriate reasoning actions before answering questions. By defining atomic reasoning action modules, searching for optimal action trajectories, and training LLMs to plan for reasoning questions, we enable LLMs to dynamically adapt to specific questions and their inherent capability. The flexibility of our two learning paradigms, i.e., external and internalized planner tuning, further highlights the adaptability of our method to different LLMs. Our experimental results show the effectiveness of \method, revealing the promise of harnessing explorations and evaluations to turn LLMs into planners for better reasoning.
% In the experiment, we show consistent improvements across a variety of reasoning tasks and settings. 
% Our work showcases the potential of harnessing explorations and evaluations to improve reasoning performance by making the LLM a smarter reasoning action planner.
% Future work will focus on exploring the automated construction of atomized action modules and scaling up the training data to achieve better results.

\bibliography{iclr2025_conference}
\bibliographystyle{iclr2025_conference}

\appendix
\newpage
% \section{Atomic Reasoning Modules}
% \label{Appendix: atomic modules}
% \input{tables/modules}

% \section{Dataset Details}
% \label{Appendix: dataset}

% \section{Training Implementation}
% \label{Appendix: training}
% \input{figures/intro_v2}

% After searching, for the external planner, we have 1722 training pairs filtered for GPT-4o-mini as the solver LLM. For LLaMA-3-70B-instruct as the solver LLM, we have 1624 training pairs. 
% For the internal planner, our training dataset consists of 2140 pairs.
% For the baselines, we do not filter the training data.

% We fine-tune the llama-8b-Instruct model using the litgpt~\citep{litgpt-2023} library. We use a learning rate of 2e-5, the global batch size is set to 64, and the maximum
% sequence length is 4096. The models are trained for 4 epochs. 
% The models are trained with 4 A100 GPUs.

\section{Training Implementation}
\label{Appendix: implementation}
In our optimal trajectory search, we set the number of iterations ($K$) to 2 and the number of evaluation times ($N_{eval}$) to 4. The number of paths retained $N_{1}$ is set to 8 and $N_{2}$ to 3. Throughout the search, we maintain a sampling temperature of 0.4. 
Searching on the training datasets eventually yields 1722 for GPT-4o-mini, 1624 for Llama-3-70B-Instruct, and 2140 for Llama-3-8B-Instruct training examples for planner tuning, respectively.
We up-sample few-shot examples by a factor of 8 and fine-tune the planner LLM using the LitGPT library~\citep{litgpt-2023}, applying a learning rate of 2e-5, a global batch size of 64, a maximum sequence length of 4096, and training epoch of 4.

\section{Case Study}
\label{Appendix: case}
In this section, $Q_I$ is a numerical reasoning problem that can be solved with Python code easily, so our tuned LLM selected PoT to solve it. $Q_{II}$ is a question widely tested in the community where GPT-4o even tends to make mistakes, but after tuning with searched trajectory data, the LLM actively chose to use code to solve the problem. 

% \subsection{Q1: A Question from MATH datasets requiring Program-solving}
\newpage
\begin{tcolorbox}[colback=gray!10, colframe=black, title=Question 1 Case Study, label={case:Q1}]
\textbf{Question I:}

A book with 50 pages numbered 1 through 50 has its pages renumbered in reverse, from 50 to 1. For how many pages do both sets of page numbers share the same digit?
\\
\\
\textbf{Answer from GPT-4o-2024-08-06 (Incorrect):}

To solve this problem, we need to determine how many pages have the same ones digit in both the original numbering (1 to 50) and the reversed numbering (50 to 1).

First, let's understand the problem: 
- The original numbering of the pages is from 1 to 50.
- The reversed numbering is from 50 to 1, meaning page 1 becomes 50, page 2 becomes 49, and so on, until page 50 becomes 1.
We need to find pages where the ones digit of the original page number is the same as the ones digit of the reversed page number.
Let's denote the original page number as \( n \) and the reversed page number as \( 51 - n \). We are looking for \( n \) such that the ones digit of \( n \) is equal to the ones digit of \( 51 - n \).

The ones digit of a number is the remainder when the number is divided by 10. Therefore, we need:

\[
n \equiv 51 - n \pmod{10}
\]

Since \( 51 \equiv 1 \pmod{10} \), we have:

\[
2n \equiv 1 \pmod{10}
\]

Thus, multiplying both sides of the congruence by 5 gives:

\[
n \equiv 5 \times 1 \equiv 5 \pmod{10}
\]

This means \( n \) must have a ones digit of 5. The pages with a ones digit of 5 in the original numbering are 5, 15, 25, 35, and 45.

Thus, the number of pages that share the same one's digit in both numbering systems is: 5\\ 

\textbf{Answer from our internal planner tuned Llama3-8b-Instruct:}

It can efficiently iterate through the range of page numbers to count occurrences of the same ones digit. Programming solver provides a quick and reliable method to handle the counting of digits across a defined range. \newline Required Action: programming 
\begin{lstlisting}[language=Python]
count = 0
for original_page in range(1, 51):
    renumbered_page = 51-original_page 
    # Check if the ones digit is the same 
    if original_page % 10 == renumbered_page % 10:
        count += 1
print(count) 
\end{lstlisting}
After execution, we get: 0 
\newline Answer: 0

\end{tcolorbox}

% \subsection{Q2: A simple Question from wild web}
\begin{tcolorbox}[colback=gray!10, colframe=black, title=Question 2 Case Study, label={case:Q2}]
\textbf{Question II:}

How many r in word `strawberry'?
\\
\\
\textbf{Answer from GPT-4o-2024-08-06 (Incorrect):}

The word ``strawberry" contains 2 letter ``r"s.\\ \\
\textbf{Answer from our internal planner tuned Llama-8b-Instruct:}

It can efficiently count the occurrences of a specific character (in this case, `r') in a word. Programming\_solver provides a quick and reliable method for counting characters in text, ensuring accuracy in the result. \newline Required Action: programming
\begin{lstlisting}[language=Python]
word = `strawberry' 
count=word.lower().count(`r') 
print(count)
\end{lstlisting}
After execution, we get: 3 
\newline Answer: 3

\end{tcolorbox}

% \subsection{Q3: A Question using actions from Analysis layer}
% \input{case/Q3}

% \input{tables/case}

\section{Sub-Task Performance of BBH}
\label{Appendix: bbh}
\begin{table}[]
\centering
\begin{tabular}{llll}
\hline
Sub Task                                 & CoT           & PoT   & \method     \\ \hline
object counting                          & 94.4          & 98.0  & \textbf{98.0}  \\
temporal sequences                       & 96.8          & 83.6  & \textbf{97.2}  \\
tracking shuffled objects five   objects & 95.2          & 78.8  & \textbf{96.4}  \\
word sorting                             & 50.8          & 98.0  & \textbf{98.0}  \\
salient translation error   detection    & 61.6          & 59.2  & \textbf{62.6}  \\
penguins in a table                      & 71.2          & 54.8  & \textbf{73.3}  \\
snarks                                   & 83.1          & 69.1  & \textbf{83.3}  \\
ruin names                               & 81.2          & 80.2  & \textbf{81.2}  \\
web of lies                              & 91.2          & 59.2  & \textbf{92.0}  \\
navigate                                 & 96.4          & 90.4  & \textbf{96.8}  \\
date understanding                       & \textbf{80.8} & 52.8  & 76.4           \\
hyperbaton                               & \textbf{92.0} & 67.6  & 91.0           \\
dyck languages                           & 38.0          & 42.4  & \textbf{54.4}  \\
tracking shuffled objects three objects  & 99.2          & 89.2  & \textbf{99.2}  \\
formal fallacies                         & 82.0          & 74.4  & \textbf{82.0}  \\
tracking shuffled objects seven objects  & 91.2          & 73.2  & \textbf{92.8}  \\
causal judgement                         & \textbf{62.6} & 62.6  & 62.0           \\
sports understanding                     & 85.6          & 74.8  & \textbf{87.6}  \\
logical deduction five objects           & 85.6          & 80.4  & \textbf{88.8}  \\
movie recommendation                     & 62.0          & 56.0  & \textbf{62.8}  \\
logical deduction three   objects        & 99.2          & 96.4  & \textbf{99.2}  \\
multistep arithmetic two                 & 98.0          & 100.0 & \textbf{100.0} \\
boolean expressions                      & 99.2          & 96.8  & \textbf{99.2}  \\
geometric shapes                         & 56.4          & 65.6  & \textbf{78.8}  \\
disambiguation qa                        & 42.8          & 40.8  & \textbf{45.6}  \\
logical deduction seven   objects        & 76.8          & 71.6  & \textbf{82.0}  \\
reasoning about colored   objects        & 87.6          & 66.0  & \textbf{87.6}  \\ \hline
\textbf{Average}                         & 80.3          & 73.9  & \textbf{84.2}  \\ \hline
\end{tabular}
\caption{Results of BBH sub-tasks.}
\label{table: bbh}
\end{table}
Table~\ref{table: bbh} shows the results of BBH sub-tasks of GPT-4o-mini as task-solving LLM and our tuned Llama-3-8B-instruct as the planner. From the table, we can see that our planner has learned to select appropriate actions based on the task. For instance, in the word sorting task, our model consistently uses code to solve the problem. Additionally, for the Dyck languages task, our method outperforms both CoT and PoT. This is because the task is easy to verify, and our model proactively requests GPT-4o-mini to verify the answer, thereby improving performance. The only exception is the date understanding task. Upon analysis, we found that while code could solve this type of problem using Python's ``datetime'' library, it often fails to follow the required output format. This leads to lower accuracy in our method's prediction when choosing writing a program as the reasoning action.

\section{Prompts Used in Experiments}
\label{Appendix: Prompts}
\begin{tcolorbox}[colback=gray!10, colframe=black, title=Prompt for query rewrite module, label={prompt:query rewrite}]
In this step, you need to reveal the Core Question with only a simple sentence and useful information. The output follows the format:

core question:...

Note: Please extract the question-solving information related to the problem, and list them one by one.

useful information:...
\end{tcolorbox}

\begin{tcolorbox}[colback=gray!10, colframe=black, title=Prompt for query decomposition module, label={prompt:query decomposition}]
In this step, you need to reflect on the problem, and describe it in your own words. Analyze how you can decompose the problem into smaller, more manageable sub-tasks. Pay attention to small details, nuances, notes and examples in the problem description.
 \end{tcolorbox}
\begin{tcolorbox}[colback=gray!10, colframe=black, title=Prompt for CoT module, label={prompt:CoT}]
In this step, you need to think step by step with words, solve the problem and get the answer.
\end{tcolorbox}

\begin{tcolorbox}[colback=gray!10, colframe=black, title=Prompt for PoT module, label={prompt:PoT}]
In this step, you need to write Python codes to solve the query. Use the simplest and most straightforward programming methods to solve the problem. For instance, if a query can be efficiently solved using a brute force method, prefer it over heuristic or more complex methods. Utilize any available and commonly-used libraries that can simplify the task or improve code maintainability. All the calculations must leverage codes. Print out the results with the print() function. Before executing the program, you have no idea of the final answer. Don't show it in your comment or code. And don't use the plot function.

In this step, start with ``\# Now write Python codes to answer this question and use \text{print()} to print out the result''
\end{tcolorbox}

\begin{tcolorbox}[colback=gray!10, colframe=black, title=Prompt for self-verification module, label={prompt:verifier}]
In this step, you need to carefully verify the correctness of the previous thoughts with natural language. You need to formulate a verification question (not the same question as before) based on the final answer and then verify the final answer you have. If the results are incorrect, the last line should end up with ``The answer is: incorrect". Otherwise, the last line should end with ``The answer is: correct"
\end{tcolorbox}
\begin{tcolorbox}[colback=gray!10, colframe=black, title=Prompt for explanation generation, label={prompt:explanation}]
\textbf{Action Categories}:

1. Understanding process:
query rewriting: Rewrite the question and answer it.
Decomposition: Decompose the questions into multiple subtasks to solve the sub-question.
2. Solving process:
chain of thought: For step-by-step reasoning with language.
programming: For programming solver.
3. Verification process:
self-verification: To check the correctness of the solution.

\textbf{Task Instruction}: For the given question, explain why the above Required actions are necessary.

\textbf{Example 1:}

Query: {{
Find $2 \cdot 5^{{-1}} + 8 \cdot 11^{{-1}} \pmod{{56}}$. Express your answer as an integer from $0$ to $55$, inclusive.
}}

Required Action: programming, self-verification

Explanation:
This is a Modular arithmetic problem.
The problem can be solved using straightforward python code with sympy library, particularly modular arithmetic. Besides, this type of problem is relatively easy to verify. After computing the result, one can check the calculations step by step to ensure correctness and verify that the final answer is within the given range (0 to 55 inclusive). Programming solver is more efficient and accurate for this type of calculation and the verifier ensures the correctness of the result and adherence to the given constraints.

...
\textbf{(multiple examples)}

Query: Given Query

Required Action: Actions After Searching

Explanation:
\end{tcolorbox}

\end{document}